# DisorientLiDAR: Physical Attacks on LiDAR-based Localization

Yizhen Lao,[1,2,#] Yu Zhang,[1,#] Ziting Wang[1], Chengbo Wang[2], Yifei Xue[2] and Wanpeng Shao.[1,*]

[1] *College of Information Science and Engineering, Hunan University, No. 2, Lushan South Road, Yuelu District, Changsha, Hunan Province, 410082, China.*

[2] *School of Design , Hunan University, No. 2, Lushan South Road, Yuelu District, Changsha, Hunan Province, 410082, China.*

*\*Email:* wpshao@hnu.edu.cn (W. Shao)

## Abstract

Deep learning models have been shown to be susceptible to adversarial attacks with visually imperceptible perturbations. Even this poses a serious security challenge for the localization of self-driving cars, there has been very little exploration of attack on it, as most of adversarial attacks have been applied to 3D perception. In this work, we propose a novel adversarial attack framework called DisorientLiDAR targeting LiDAR-based localization. By reverse-engineering localization models (*e.g.*, feature extraction networks), adversaries can identify critical keypoints and strategically remove them, thereby disrupting LiDAR-based localization. Our proposal is first evaluated on three state-of-the-art point-cloud registration models (HRegNet, D3Feat, and GeoTransformer) using the KITTI dataset. Experimental results demonstrate that removing regions containing Top-K keypoints significantly degrades their registration accuracy. We further validate the attack's impact on the Autoware autonomous driving platform, where hiding merely a few critical regions induces noticeable localization drift. Finally, we extended our attacks to the physical world by hiding critical regions with near-infrared absorptive materials, thereby successfully replicate the attack effects observed in KITTI data. This step has been closer toward the realistic physical-world attack that demonstrate the veracity and generality of our proposal.

## Table of Contents

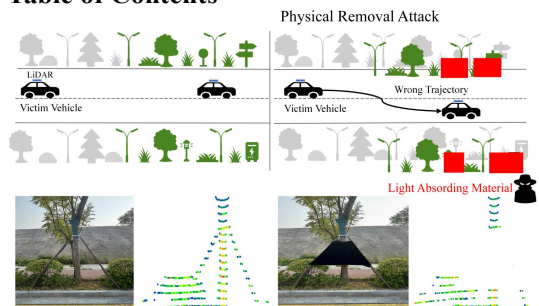

*Innovative Description:* Defeat LiDAR-based localization performance by physically hiding key regions.

## 1. Introduction

Localization is an essential task for modern autonomous vehicle (AV) systems that allows AVs to accurately determine their position and orientation in the environment. These localization systems take advantage of sensors such as LiDARs and cameras for safe navigation, decision-making, and planning of driving actions. LiDAR sensors, in particular, are used to capture depth measurements of the surroundings with high accuracy in 3D point clouds to localize vehicles. Recent end-to-end deep learning approaches (HRegNet,[1] D3Feat,[2] Geotransformer[3]) have emerged as a popular choice for LiDAR-based localization which have shown remarkable performance on localization benchmarks (KITTI,[4] ETH dataset,[5] Nuscenes dataset,[6] CarlaScenes[7]).

### 1.1 Motivations

For a safety-critical system like an AV, it is as essential for its components to be high-performing as it is for them to be reliable and robust. Despite previous research showing that AVs are vulnerable to attacks on LiDAR sensors that exploit their perception models such as obstacle detection,[8-11] obstacle hiding,[12-15] and motion prediction[16]-few existing works have rigorously evaluated the robustness of localization models against adversarial attacks. Fukunaga *et al.*,[17] misled scan matching algorithms in a simulated environment via LiDAR spoofing (laser-based point injection), but the method was not applied to real-world adversarial scenarios. Thus, we tackle the following research questions in this paper: *(i) Is it possible to remotely and stealthily cause inaccuracies in the AV's LiDAR-based localization system? (ii) How can an attacker perform such an attack under realistic conditions? (iii) What are the implications of such attacks on AV frameworks and localization models, and how is it possible to defend against them?*

**1.2 Contribution and paper organization**

To answer these research questions, we propose a new attack framework called DisorientLiDAR. We investigate the feasibility of manipulating LiDAR data acquisition to degrade the localization accuracy and robustness, which further raises safety risks for the AV itself, pedestrians, and other vehicles. Specifically, we find that, hiding key regions that serve as vital cues for point cloud registration can possibly defeat the AV's localization performance in the scene, as depicted in Fig. 1.

First, we carefully study and analyze the state-of-the-art deep neural network (DNN)-based point cloud registration approaches in section 3, which are able to handle the irregularity of point clouds and achieve high registration accuracy and can be applied as the core of the AV's localization system. However, we also observe that these data-driven DNN solutions share a common vulnerability, namely, entirely rely on detecting and matching local correspondences with significant geometric features (aka key region) between the two input point clouds to perform the registration task. This observation provides a novel insight that one can defeat the performance of point cloud registration DNN models by hindering the key region's matching.

Next, we develop an adversarial attack framework targeting AV's localization system by introducing a new threat model in section 4, specifically: i) We design a strategy that can identify the key regions among the pre-defined ambush area by estimating the confidence scores that indicates the contribution to the DNN-based registration model. ii) The attack setup is to hide the key regions from the LiDAR scanning by using an infrared light-absorbing material to cover these regions. iii) We introduce a method to systematically determine the influence of the attacks on AV's localization task. The workflow of this framework is illustrated in Fig. 2, providing a clear overview of the steps involved in the adversarial attack process.

Then, we quantify the capability of the attacker in different scenarios in section 5. To validate our approach, we examine the effectiveness of the key region hidden via synthetic experiments on the KITTI dataset by modeling the attacker's capability for different scenarios, including different numbers and sizes of removal regions. Towards generating physical attacks, we use an infrared-absorbing product to cover some common objects around the key correspondence. Such products can be used in any scene and are capable of hiding objects from a strong LiDAR detector. We demonstrate that the point cloud registration accuracy drops in all three tested models, leading to vehicle localization failure.

Finally, we summarize two defense strategies: adversarial training and anomaly detection, as detailed in section 6. To summarize, this work aims to model, measure, and demonstrate the capability of hiding the key region information by leveraging state-of-the-art point cloud registration models' mechanism and helps defend against the threat to current and future AVs.

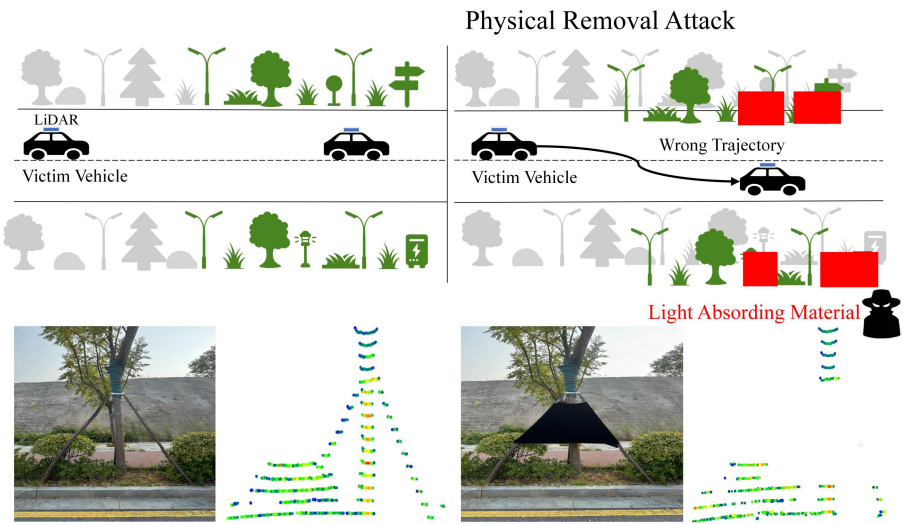

**Fig. 1:** DisorientLiDAR attack. The attack goal is to disorient the victim vehicle with erroneous localization and wrong trajectory plan (top-right) by placing several light-adsorbing materials in front of some critical objects, making them "disappear" from the LiDAR scanning (bottom-right).

Our contributions are summarized as follows:

• We identify and model a novel physical location attack DisorientLiDAR on the AV localization system that hides key regions along the street and leads to a decrease in point cloud registration accuracy. We also propose a strategy that can achieve the real physical attack in real vehicle scenarios.

We evaluate the attack impact on three SOTA point cloud registration models. We also model the attacker capability, challenges, and limits of DisorientLiDAR on a commercial AV perception system.

• We validate our findings by showing consequences for autonomous vehicles on the production-grade AV simulator and conducting real-world attacks on vehicles. Finally, we propose two enhanced strategies to mitigate the threat.

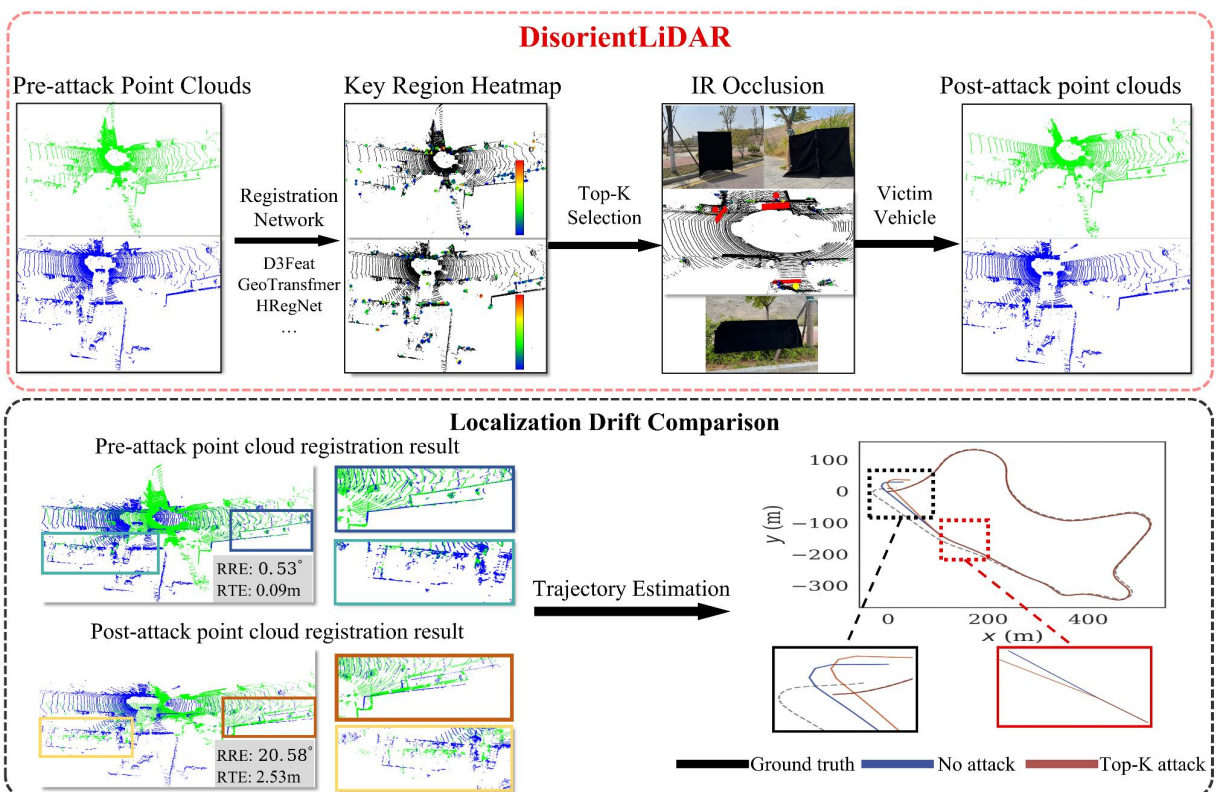

**Fig. 2:** The workflow of the DisorientLiDAR.

## 2. Related work

### 2.1 Lidar-based registration in AVs

Existing point cloud registration can be grouped into two categories, namely scan-to-scan and scan-to-HD map (High-definition map):

Scan-to-scan point clouds registration. Previous scan-to-scan point cloud registration networks have relied on traditional algorithms such as 3D-NDT,[18] 4PCS,[19] SAC_IA,[20] and KISS_ICP[21]. However, these methods face limitations due to their sensitivity to initial conditions, vulnerability to noise, and high computational demands, which impact their efficiency and accuracy in complex scenarios. As PointNet[22] emerges as the pioneering 3D deep learning network capable of directly extracting features from input point clouds, researchers began to apply this 3D deep learning technique to point cloud registration.[23] The main pipeline of such registration models involves using a 3D DNN to extract features for correspondence estimation. Then, a single-step optimization (*e.g.,* SVD) can be used to estimate the transformation matrix based on the estimated correspondences. These models can be roughly classified into two classes: keypoint-based and keypoint-free. Keypoint-based models heavily rely on detected significant keypoints for transformation estimation, such as those referenced in HRegNet[1], D3Feat[2] and Deep global registration[24]. In contrast, keypoint-free methods consider all potential correspondences rather than relying on several critical points, as seen in GeoTransformer[3] and RDMNet[25]. In our experiments, we will conduct attacks on these two types of registration networks and then analyze the influence of the attack parameters on their registration results.

Scan-to-HD map registration. The typical approach for HD map registration involves registration between newly scanned frames and a local map. This method often requires an initial pose to find the closest local map at first. The registration process is similar to scan-to-scan registration, where feature points need to be extracted from the scanned data and local map. Then, correspondences are established by these feature points' descriptors. Once sufficient correspondences are found, geometric transformations are computed to achieve registration. Some researchers apply this approach to estimate the final pose of the vehicle by aligning current sensor readings (raw laser scans or visual features) with pre-built map.[26,27] The others focus on updating HD maps, since the "old" maps collected in the past may have deviations or changes over time. The registration of newly collected point clouds on the HD map ensures that the map information remains consistent with the real-time environment.[28,29]

### 2.2 Adversarial attacks on LiDAR

Most existing studies focus on adversarial attacks against the perception module in AV.[30] These attacks can be categorized into three types: signal injection, object removal, and object mis-categorization. In signal injection attack,[31-35] the adversary can inject additional point clouds into the scene to generate a non-existent object by shining a laser toward the victim sensor. Since they share the same physical channels, the victim sensor accepts these malicious signals as legitimate. Object removal, on the other hand, can be achieved by placing physical objects on the target vehicle.[36,37] These objects are designed with specific shapes and sizes to reduce their confidence score in the victim AV's detection models that make the host vehicle undetectable. In a mis-categorization attack, the adversary can compromise the victim's classifier by injecting a backdoor, which causes the self-driving car to misinterpret certain objects as obstacles.[38] Moreover, some attackers place objects at specific locations to mislead the LiDAR perception module of an autonomous vehicle, thereby misleading the trajectory prediction module.[39]

### 2.3 Adversarial attacks on AV's localization

Currently, research on adversarial attacks against localization systems is scarce. Fukunaga *et al.*,[17] investigated LiDAR spoofing attacks by injecting fake point clouds using a laser device. However, this method is constrained by both its localized spatial manipulation (limited to specific regions) and hardware- imposed restrictions on the number of fake points injected per scan. Unlike Fukunaga *et al.* [17], P. Kumar *et al.*,[40] process a newly collected LiDAR scan to produce a new frame including adversarial points produced by a generative adversarial module. The attack is designed to cheat the Google Cartographer, a SLAM-based algorithm, leading to incorrect pose estimation and map construction. The key limitation is the attackers require full access to both the vehicle's localization model and the LiDAR system—a highly privileged level of authorization that is extremely difficult to obtain in practice. Yoshida *et al.*,[41] developed a LiDAR spoofing method by manipulating LiDAR scan data through controlling laser emission timing and direction to deceive ICP[42] and NDT,[18] which induces pose estimation errors and distorted map reconstruction. However, like P. Kumar *et al.*,[40] this method also requires physical access to the LiDAR hardware, which is nearly infeasible in real-world scenarios. Unlike these methods, our attack requires no physical access to LiDAR sensors but can achieve the unexpectedly high attack success rate by strategically deploying stealthy disruptions through key-region hidden manipulation. Furthermore, our method is evaluated on real-world autonomous driving systems (*e.g.*, using benchmark datasets like KITTI or real vehicle- collected point clouds).

## 3. Attack goal and threat model
### 3.1 Attack goal
To analyze potential attacks on registration networks, we first review how point cloud registration is performed. Given two point clouds, $P = \{p_i \in \mathbb{R}^3 | i = 1, ...N\}$, and $Q = \{q_i \in \mathbb{R}^3 | i = 1, ...M\}$, point clouds registration aims to estimate the rigid transformation $T = \{R, t\}$ that aligns them, where $R \in SO(3)$ is the rotation matrix and $t \in \mathbb{R}^3$ is a translation vector, The optimal transformation $R^*, t^*$ is solved by:

$$R^*, t^* = \underset{R,t}{\operatorname{argmin}} \sum_{(p^*_{xi}, q^*_{xi}) \in C^*} w_i ||Rp^*_{x_i} + t - q^*_{x_i}||^2 \qquad (1)$$

where $C^*$ represents the set of reliable point correspondences between P and Q. $w_i$ denotes the confidence score for each correspondence pair. The accuracy of the registration heavily relies on the high-quality corresponding points. Therefore, by identifying and removing a minimal yet sufficient subset of the most reliable correspondences, we can force the registration model to rely only on sub-optimal matches, thereby inducing significant alignment errors.

### 3.2 Target point clouds registration models
In this study, we focus on attacking learning-based point cloud registration. Our investigation begins with a systematic analysis of state-of-the-art 3D keypoint detectors and their registration pipelines, including:

HRegNet[1] employs a hierarchical framework for point cloud registration by leveraging multi-scale keypoint extraction and dual-consensus matching. For keypoint detection, the method hierarchically downsamples raw point clouds to identify sparse yet geometrically stable keypoints with discriminative descriptors. For registration, it then establishes reliable correspondences through a dual-consensus mechanism: bidirectional reciprocity enforces mutual nearest neighbor relationships in feature space, while neighborhood consistency verifies spatial distribution coherence across local regions. Finally, the optimal transformation is computed via weighted SVD, where weights are confidence scores predicted by the network to prioritize high-quality matches.

D3Feat[2] formulates point cloud registration as a joint optimization of keypoint detection and feature description. For keypoints detection, it employs a fully convolutional U-Net like architecture to extract multi-scale geometric features and detects keypoints by identifying local maxima in feature maps via the non-maximum suppression mechanism, prioritizing points with high repeatability and distinctiveness. For registration, RANSAC is used to estimate transformation matrices.

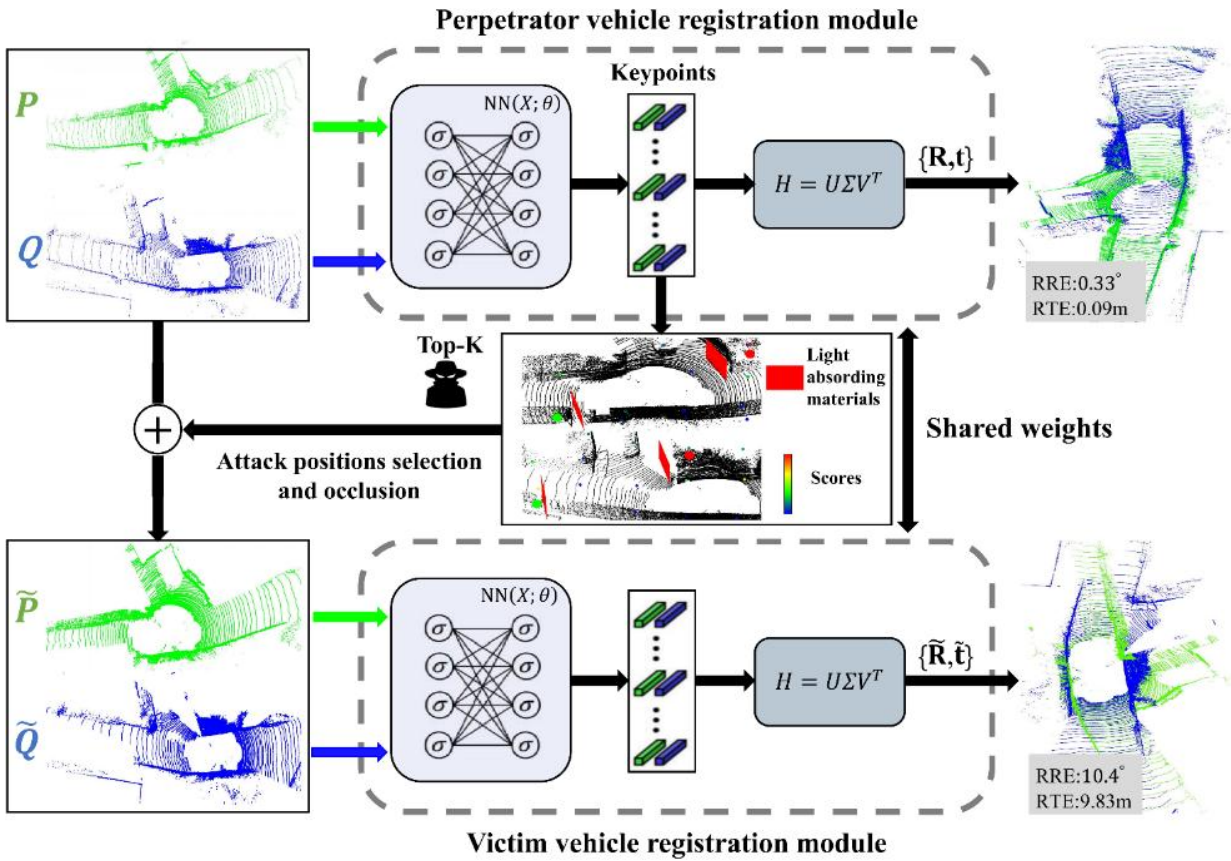

**Fig. 3:** Overview of the DisorientLiDAR attack pipeline. The attackers first extract high-confidence keypoints from the raw point clouds (P, Q) using an identical copy of the victim vehicle's localization model. The attacker then selects the top K most salient keypoints and hide the corresponding regions using near-infrared-absorbing materials. The hidden causes the victim vehicle's LiDAR to miss these critical geometric regions, ultimately degrading registration performance.

Geotransformer[3] integrates geometric invariants encoding with a local-global registration (LGR) module for fast and accurate registration. In the detection phase, it employs a KPConv-FPN architecture to extract multi-scale geometric features from point clouds hierarchically. Additionally, it adopts a geometric transformer module to learn transformation-invariant geometric representations. Specifically, the geometric transformer encodes intra-point-cloud geometric structures and inter-point-cloud geometric consistency to extract high-quality superpoint correspondences. These superpoint matches are subsequently used to compute dense local point correspondences through an optimal transport layer. During the local registration stage, a transformation matrix $T = \{R, t\}$ is computed for each superpoint region using the local dense point correspondences. During the global registration stage, the transformation $T_i = \{R_i, t_i\}$ with the most inlier matches is selected, then the final
transformation matrix is re-estimated by N iterations based on surviving inlier matches.

### 3.3 Previous knowledge
For the model attack, we assume that the adversary can learn the behavior of the victim registration models by reading publicly available documents (*e.g.*, manuals, datasheets, and open-source code). For attacks on open-source AV platforms, we also assume that attackers should have access to the platform code, data, and localization algorithms of Robot Operating System (ROS)-based systems (such as Autoware) or Apollo-based autonomous driving frameworks, to estimate the spoofing regions of the target AV. This assumption is less restrictive than prior adversarial attacks on LiDAR. Furthermore, ROS-based and Apollo-based AV frameworks are widely used in the AV industry and provide a basis for the attacker to understand the features of common systems and potential vulnerabilities.

### 3.4 Attack scenarios
After acquiring the victim's driving route and localization algorithm specifications, adversaries strategically select a target road segment—typically one with minimal pedestrian traffic where the victim vehicle will pass through. In more detail, attackers deploy near-infrared (NIR) absorbing materials (*e.g.,* specialized fabrics or coatings) to hide pre-selected key regions on both sides of the road, then wait for the victim to drive by. In AV settings, these attacks are stealthy.

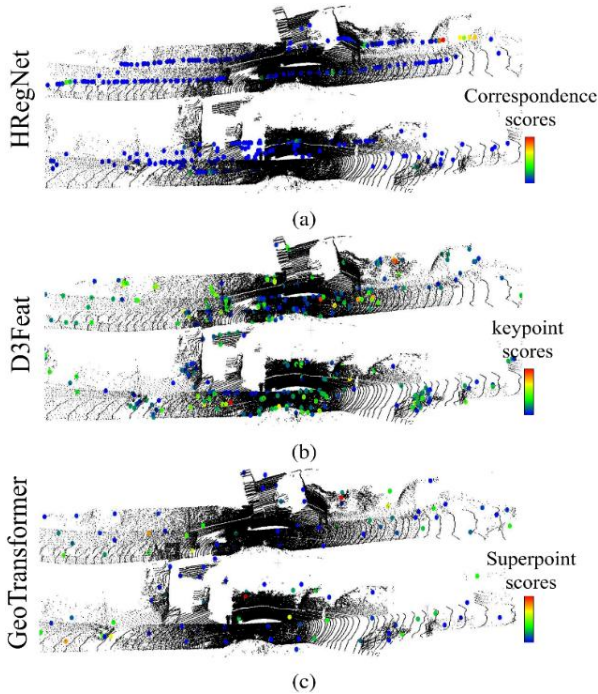

**Fig. 4:** Visualization of keypoints in an input pair. Due to variations in confidence estimation mechanism, scores are normalized to 0 to 1 based on $(S_i-S_{min})/(S_{max}-S_{min})$. (a) Scores reflect the consensus-verified correspondence weights. (b) Scores indicate the joint confidence of density-invariant saliency and feature channel significance. (c) Scores represent the confidence of superpoint matches weighted by global inlier counts.

## 4. DisorientLiDAR attack design
### 4.1 Registration saliency map
Fig. 3 illustrates our DisorientLiDAR framework. The attack effectiveness varies across different registration architectures, so the key step is identifying target model-specific keypoints that significantly impact registration performance, where:

• For HRegNet, it adopts three-stage cascaded registration with {256, 512, 1024} point correspondences. The initial 256-point matches significantly impact the final transformation accuracy, since the first-stage output (R, t) propagates through subsequent stages. This hierarchical architecture makes the first layer the optimal attack target: 256 correspondences in the first layer shown in Fig. 4(a) are evenly distributed in the input dataset with only a few points having high confidence scores, which are particularly vulnerable to strategic perturbations.

• For D3Feat, only keypoints with the highest confidence scores are selected for RANSAC registration. Therefore, removal of the K most confident points (as ranked by the network's prediction) can degrade the quality of keypoints used for RANSAC. As shown in Fig. 4(b), we observe that most of high-confidence keypoints are concentrated in a few well-localized regions. Such a concentration of critical features makes the algorithm particularly susceptible to key region hidden attacks.

• For GeoTransformer, its registration performance critically depends on identifying high-consensus superpoint matches that yield transformations with maximal inlier counts. As analyzed in section 3.2, removal of these key regions containing optimal candidates forces the model to rely on suboptimal superpoints with fewer inlier matches.

### 4.2 Key region hidden strategy
Our hypothesis is that strategically removing K critical geometric regions with the highest contribution scores (denoted Top-K) will significantly degrade registration performance. To explore this concept further, we also tried several other region-hiding strategies, including hiding regions around randomly selected K positions (denoted Rand-K) and hiding regions around K positions with the minimum contribution (denoted Min-K).

However, due to the limitations of algorithms, some wrong matches extracted by the registration model in the reference scan and source scan do not correspond to the same location in the real-world scene. Even one key point is matched by multiple keypoints,[3] leading to inconsistency with the physical attacks. In order to ensure that the selected object in the reference scan and the source scan point at the same position in simulated experiments for these ambiguous matches, we select the keypoints in the source scan and transform them to the reference scan using ground truth transformation.

It is noted that some selected points' positions are unsuitable for physical attacks due to either detection risks or accessibility constraints. Specifically, some points in the middle of the road are noticeable, and some points on top of ground objects (*e.g.,* on road signs or tree canopies) are hard to reach, making attack operations infeasible. Therefore, we propose a four-stage screening process for attack feasibility:

• Height Filtering: We remove ground points using cloth simulation filtering method,[43] and then exclude the point number greater than 3 m above ground as physically unreachable.

• Trajectory Proximity Check: To maintain stealth, we exclude attack points that intersect the vehicle's navigation path to prevent detectable anomalies in driving behavior.

• Overlapping Handling: For multiple keypoints belong to the same object or direction relative to the vehicle, we keep only the nearest one to the vehicle.

• Ambush tool placement: To ensure full coverage, we place the ambush tool 1 m ahead of the target point (perpendicular to the ground).

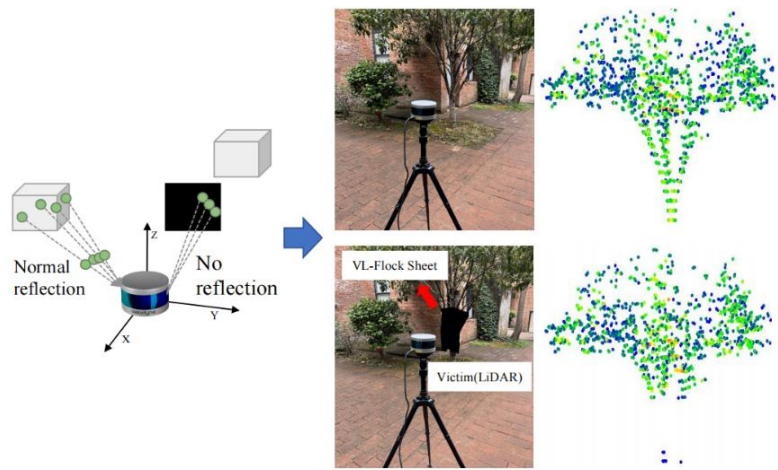

**Fig. 5:** Comparison of scanned objects with and without VL Flock Sheet coverage.

### 4.3 Physical implementation

To simulate a physical attack in the real world and make attacks less detectable, we adopt a simple method: Covering the target with a black cloth. The cloth is capable of absorbing the infrared light (905 nm) emitted by the LiDAR. In our study, we selected a product called VL Flock Sheet produced by KOYO CO., LTD.,[a] which achieves > 95% light absorption in the range λ < 2200 nm (NIR-SWIR). As shown in Fig. 5, the top image depicts the original scanned object, while the bottom image shows the object covered by black cloth. Almost no point cloud data is observed in the covered regions.

### 5. Evaluations and results

In section. 5.1, we first evaluate simulated DisorientLiDAR attacks on the KITTI dataset,[4] and analyze how attack parameters (*e.g.*, ambush positions and tool sizes) affect scan-to-scan registration and trajectory estimation in HRegNet,[b] D3Feat,[c] GeoTransformer.[d] Then, we validate simulated attacks on the localization module of the commercial AV platform Autoware in section.5.2. Finally in section.5.3, we deploy DisorientLiDAR attack in real-world scenarios.

### 5.1. Evaluation of simulated attack on KITTI dataset
#### 5.1.1 Experimental setup

The KITTI odometry dataset is collected by a Velodyne HDL64 LiDAR in Germany. KITTI provides LiDAR scans collected in diverse environments with the corresponding ground-truth poses. In KITTI odometry, only 00-10 contains the ground truth pose, so the KITTI odometry is normally split into three sets: 00-05 for training, 06-07 for validation, and 08-10 for testing in registration models. Our attack experiments target the testing phase, so we directly performed our attack on the 08-10 dataset.

**Parameter study:** To evaluate the impact of incremental perturbations on registration performance, we varied the number of ambush positions from 1 to 10. For each number of ambush positions, we set the cloth's side length to incremental values: 0.3, 0.6, 0.9, 1.2, 1.5, 1.8, 2.1, 2.4, 2.7, and 3m.

**Metrics:** In our study, we employ the same evaluation metrics as those used for assessing point cloud registration networks, focusing on three key metrics: (1) Relative Rotation Error (RRE), the geodesic distance between the estimated and ground-truth rotation matrices. (2) Relative Translation Error (RTE), the Euclidean distance between the estimated and ground-truth translation vector. (3) Registration Recall (RR), the ratio of point cloud pairs for which RRE and RTE below specified thresholds. In registration studies, the averages of RRE and RTE are calculated only for the samples below the thresholds for RTE and RRE. In contrast, our approach computes the average RRE and RTE across all samples. Given that attacks typically do not result in substantial decrease in RRE and RTE for most samples, we set the RR thresholds at 0.5 degrees for RRE and 0.3 meters for RTE to better observe the impact of the attack on registration performance.

### 5.1.2 Result and analysis

As shown in Fig. 6, across all attack types, the Top-K attack is the most effective due to the removal of the most critical key regions, while Min-K and Rand-K attacks demonstrate inferior performance. With the same key region removal strategy, HRegNet exhibits significant performance fluctuations; in contrast, GeoTransformer shows the highest robustness. As for the trajectory estimation, compared with rand-K and Min-K, Top-K attack induces significant distortions (some routes even present an approximately $180°$ flipped). Although attack-induced perturbations appear subtle initially, mapping direction deviations become increasingly amplified as the vehicle travels further from the attack place. The detailed analysis is as follows:

**HRegNet:** The attack performance exhibits significant instability. This stems from the overlap between Top-K and Min-K correspondence regions computed by HRegNet. Consequently, removing Top-K regions (containing high-saliency correspondences) inadvertently eliminates Min-K correspondences as well. As a result in Fig. 6(a1), RTE (within the range of 0.319 m to 0.396 m) and RRE (within the range of $2.519°$ and $3.303°$) values exhibit a random pattern, accompanied by a decrease in RR (from 0.663 to 0.526). As for Min-K attack, this joint removal leads to an unexpected increase in RR.

**D3Feat:** The attack effect becomes more severe as more critical regions are removed. The reason is the 250 keypoints for each frame pair are selected by the network rather than being downsampled, so removal of each key point will lead to degradation of the registration. As shown in Fig. 6(b1): the results of the Top-K attack show that there is a consistent increase in RTE (from 0.104 m to 0.541 m) and RRE (from $0.427°$ to $2.382°$), accompanied by a decrease in RR (from 0.729 to 0.569). Besides, Min-K attacks demonstrate superior performance over Rank-K due to removal of uncertain number of keypoints.

**GeoTransformer:** It exhibits more adversarial attack resistant than D3Feat and HRegNet. Even multiple key regions are removed; it still maintains robust performance. Therefore, only under Top-K attacks with aggressive parameters, GeoTransformer becomes vulnerable. As shown in Fig. 6(c1), there is a dramatic rise in RTE (from 0.068 m to 0.677 m) and RRE (from $0.233°$ to $4.836°$), and RR falls sharply from 0.944 to 0.789.

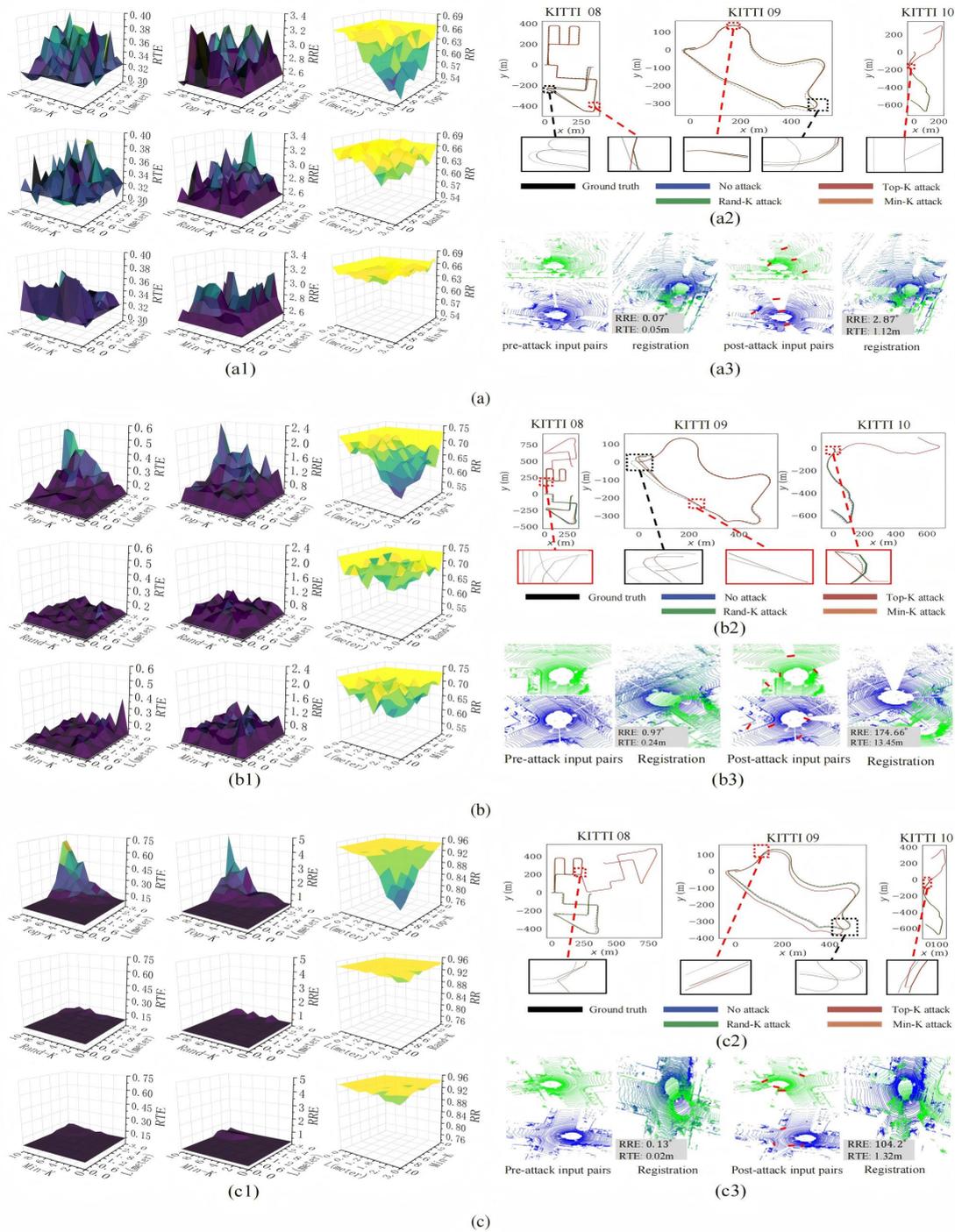

**Fig. 6:** Evaluations of region removal attacks on (a) HRegNet, (b) D3Feat, and (c) GeoTransformer with KITTI Dataset. (a1-c1) Top-K/Rand-K/Min-K attack strategies comparison; (a2-c2) Trajectory estimation under Top-K attack, it is noted that the HRegNet use every 10-frames data. (a3-c3) Attack scenario on Route 08.

### 5.2. Evaluation of simulated attack on commercial AV platform
#### 5.2.1 Experimental setup
Currently, deep learning techniques have not been applied in the localization module of AV platforms. We take the Autoware localization system as an example, it uses the Normal Distributions Transform (NDT) algorithm for localization. Unlike registration algorithms based on corresponding points, the NDT algorithm does not require finding matching points. Instead, it divides the reference point cloud into regular grids, computes the probability distribution of points within each grid (usually assumed to be a normal distribution), and then uses optimization algorithms to find the transformation matrix that best aligns the target point cloud with the reference point cloud.

In the Autoware[(1)] localization system, the vehicle achieves localization by registering real-time scanned point cloud data with a HD map point cloud. At the same time, GPS and IMU provide initial pose information for the system. The vehicle in the platform was equipped with a 128-beam LiDAR (Velodyne VLS-128). The specific attack process was as follows: We chose a crossroads as an ambush site and remove 6 distinct regions. As shown in Fig. 7(b), compared to the pre-attack point cloud, six well-defined regions are removed.

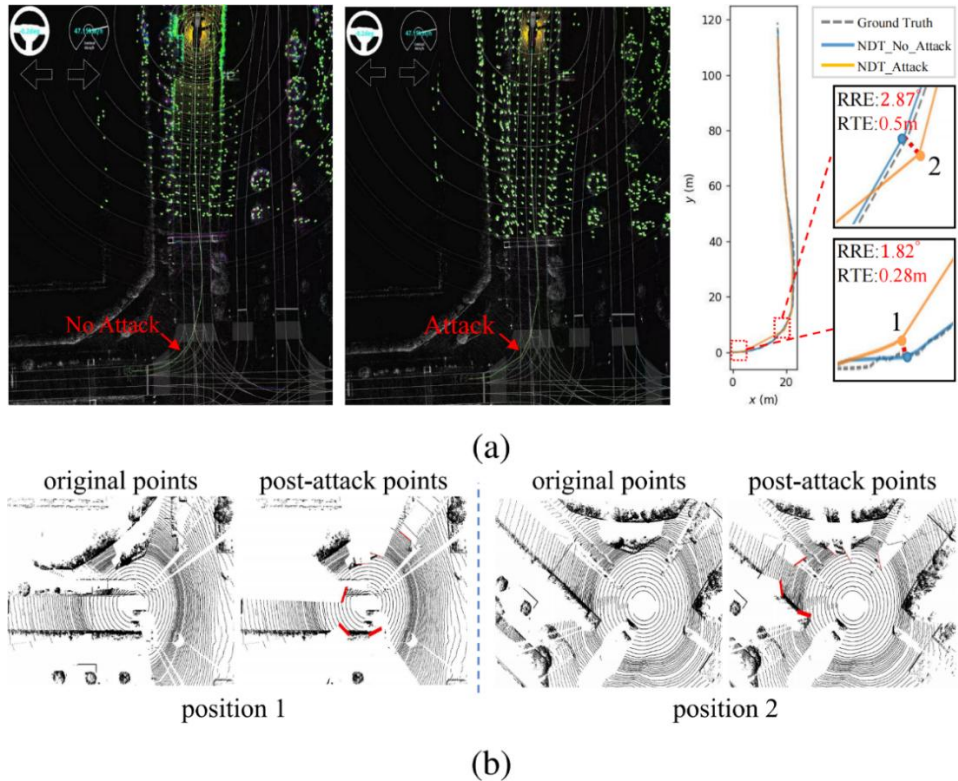

**Fig. 7:** Attack on the localization system of the Autoware platform. (a) In the two figures on the left, the grey points in the Autoware platform represent the HD map point clouds, while the green points are the point clouds collected in real - time. The positioning process involves using the initial pose provided by GPS and IMU in combination with the real-time collected point clouds. By registering the green points with the HD map, real - time coordinate information can be obtained. The green line indicates the vehicle's driving trajectory. The figures on the right show a comparison of the trajectory information before and after the attack. (b) Comparison of the point clouds collected in real-time before-and-after attack at position 1 and position 2. The red lines indicate the ambush sites.

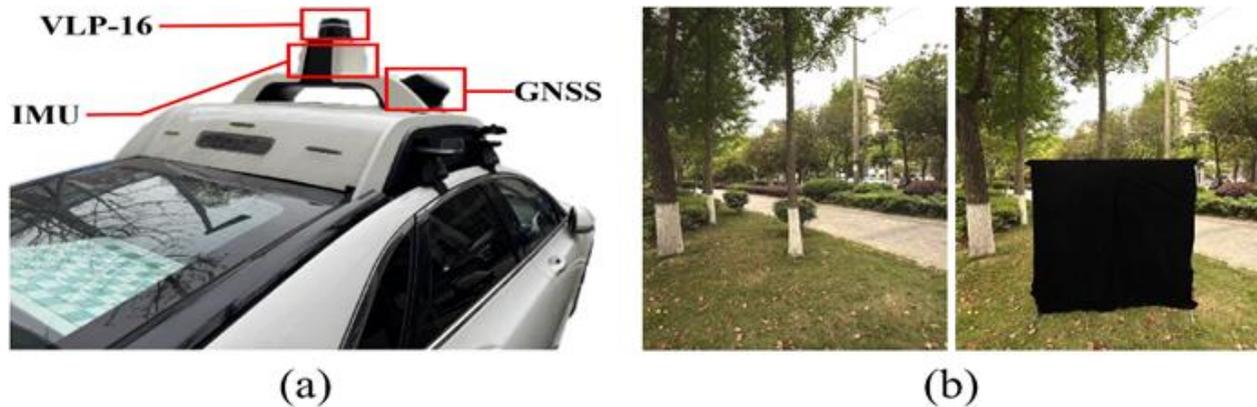

**Fig. 8:** The experiment setup. (a) AV platform (b) Light-absorbing material deployment: Original scene (left), Region hidden scene (right).

### 5.2.2 Result and analysis

As shown in Fig. 7(a), at position 1, compared to the benign case, the localization errors after the attack are as follows: RTE reached $0.28\text{m}$, and RRE reached $1.82°$. Subsequently, the vehicle continued to travel a certain distance with no registration performed. Upon reaching position 2, where the second real-time point cloud registration and localization were conducted, RTE caused by the attack, compared to before the attack, reached 0.5 m and RRE increased to $2.87°$. After the vehicle exited the attack-affected area, the trajectory was gradually corrected back to the normal path with GPS support. This result fully demonstrates that even the AV system is equipped with high-performance LiDAR and is supported by the GPS and IMU, our attack method can still lead to localization errors.

### 5.3 Evaluation of real physical attack
#### 5.3.1 Experimental setup
In this experiment, we conduct physically realizable adversarial attacks. We directly apply the models trained on the KITTI odometry to the self-recorded dataset. Fig. 8(a) shows our own platform equipped with a Velodyne VLP-16 LiDAR,[(1)] an inertial measurement unit (Xsens MTi-300), and a GNSS (INS CGI-410). We build our dataset in a road environment with ground-truth poses calculated by combining the GNSS and IMU with the state-of-the-art LiDAR SLAM method.[44] Note that the sensors, environments, and platform setups are different from the KITTI odometry datasets to others, which thoroughly tests the generalization ability of the approaches. Fig. 8(b) shows our region-hiding deployment in real-world scenarios. As explained in section 3.4, we execute physical attacks as follows: We simulate GPS-denied urban environments where AV vehicle localization relies solely on LiDAR registration models. At first, we regard our platform as perpetrator vehicle and collect reference LiDAR data and ground truth trajectory using our platform. Subsequently, we perform pairwise registration for each target model (HRegNet/D3Feat/GeoTransformer) and produce attacker trajectory through chained transformations. Then, we identify a vulnerable frame where registration errors exceed predefined thresholds (RRE $> 15°$ or RTE >2.0 m) with minimal physical occlusions. Next, we deploy our NIR-absorbing patches on selected keypoints. Finally, we regard the platform as a victim vehicle and repeat the trajectory to collect the attacked point clouds and compute the post-attacked trajectories through identical registration models.

#### 5.3.2 Results and analysis
As shown in Fig. 9(a), HRegNet performs poorly in 16-beam LiDAR registration tasks due to insufficient point cloud density for its feature extraction pipeline, where even pre-attack trajectories exhibit severe distortion compared to its ground truth trajectory. However, we can still observe the victim vehicle's registered trajectory exhibits greater deviation. The adversarial frame pairs exhibits significantly increased relative pose errors: from RRE $5.54°$ and RTE 1.1 m (against to perpetrator ground truth) to RRE $12.48°$ and RTE 2.3 m (against to victim ground truth).

Unlike HRegNet, D3Feat performs much better in 16-beam LiDAR registration task in Fig. 9(b). For the selected frame pairs under attack, the victim vehicle's registered pose exhibits substantial deviation from its ground truth trajectory.

As shown in Fig. 9(c), GeoTransformer achieves the most accurate registration performance on 16-beam LiDAR data. However, it is unexpected that only one key region hidden in selected frame pairs causes severe registration failures, especially for the RTE: 0.13 m (pre-attack, against perpetrator ground truth) increases to 5.95 m (post-attack, against to victim ground truth). In such case, the estimated poses between the selected frame pairs remain nearly identical despite actual vehicle movement. This results in significant trajectory drift, particularly at turning maneuvers.

### 6. Discussion
#### 6.1 Attack robustness analysis
In simulated environments like the KITTI dataset experiments, attackers can precisely place them at the ambush site, facing the vehicle in the source frame, which maximizing coverage of high-contribution areas. However, in real-world deployment, even if the attacker knows the general direction in which the target vehicle is moving, the material still cannot be precisely oriented to match the vehicle's exact position in the localization frame—due to uncertainty in the vehicle's trajectory and the complexity of road environments (*e.g.*, multi-lane roads, curves). This misalignment may lead to partial exposure of high-contribution regions, reducing the potential impact of the attack. These constraints make angular insensitivity a crucial robustness metric for physical attacks.

To characterize the angular tolerance of our attack, we conducted systematic experiments with seven orientations within the range of ±30°, with 10° increments. This angular range was chosen to reflect the realistic level of uncertainty an attacker may face during physical deployment in real-world scenarios, which sufficiently covers the plausible deployment deviations. The attack parameters we set are K = 5, L = 2.1 m. As shown in Fig. 10, despite significant fluctuations in RTE and RRE caused by orientation changes, the attack effectiveness degrades only modestly as orientation changes in terms of RR:

For HRegNet, RR did not exceed 59.6% under any orientation, in contrast to the 66.3% baseline performance.

For D3Feat, RR remained below 65.8% across all orientations versus 72.9% in the baseline.

For GeoTransformer, it achieved no more than 91.5% RR across all orientations compared to the baseline performance 94.4%.

This demonstrates that as long as the material covers key regions and roughly faces the oncoming direction of the target vehicle, the attack remains highly effective. The angulartolerance simplifies real-world deployment and increases thepracticality of real-world deployment.

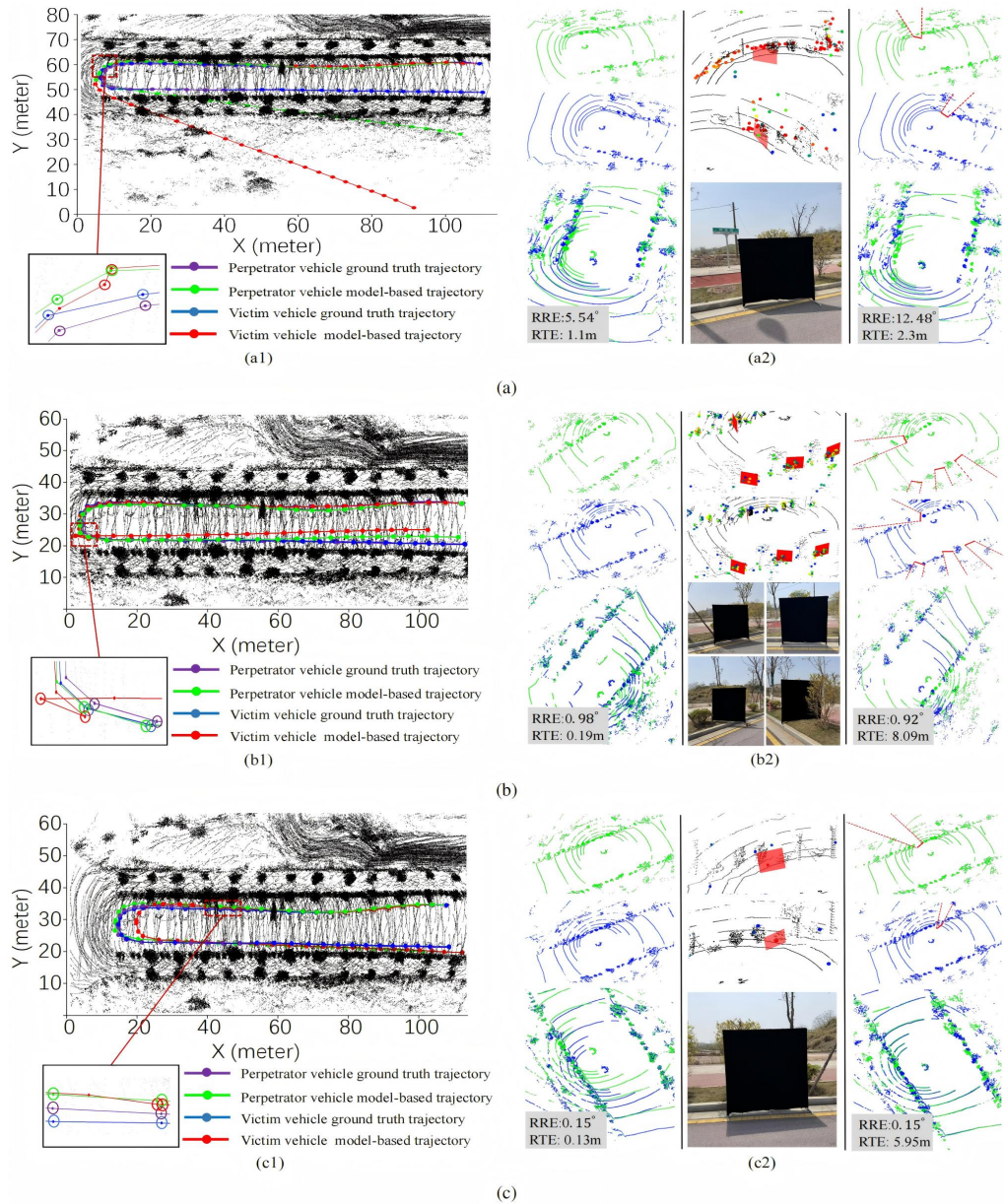

**Fig. 9:** Evaluations of physical attacks on (a) HRegNet, (b) D3Feat, and (c) GeoTransformer, respectively. (a1-c1) Trajectory comparisons between ground truth and model-predicted paths for both perpetrator and victim vehicles. (a2-c2) pre-attack input pairs and registration results (left), heat maps with selected attack positions in red rectangle and corresponding physical camouflage using VL Flock Sheet material (middle), post-attack input pairs and registration results.

## 6.2 Practical implication

Our attack method possesses following strengths: simulation-to-reality consistency, cross-sensor compatibility, no hardware modification required, and robustness to deployment angle error. These characteristics pose significant security implications for autonomous driving systems design:

- Do not rely solely on simulation:

In our experiments, we validated the attack using the KITTI dataset (simulation experiment), the Autoware platform (software), and a real-world physical environment. Across all settings, the attack effect was highly consistent. *Therefore, manufacturers must not assume that passing simulation-only tests guarantees real-world safety for their systems. Instead, they need to perform adversarial testing in physical environments to ensure that simulation results truly reflect real-world performance.* all types of LiDAR, not just a single type:

We successfully executed the attack on multiple LiDAR configurations: a 64-line LiDAR (used in the KITTI dataset in section 5.1), a 128-line LiDAR (used in Autoware in section 5.2), and a low-end 16-line LiDAR from our lab (in section 5.3). This shows that the same vulnerability exists across high-end multi-beam and cost-effective LiDAR units alike. *Therefore, manufacturers cannot harden only one particular LiDAR model; they must develop system-level defense and detection strategies that cover all LiDAR types used in their vehicles.*

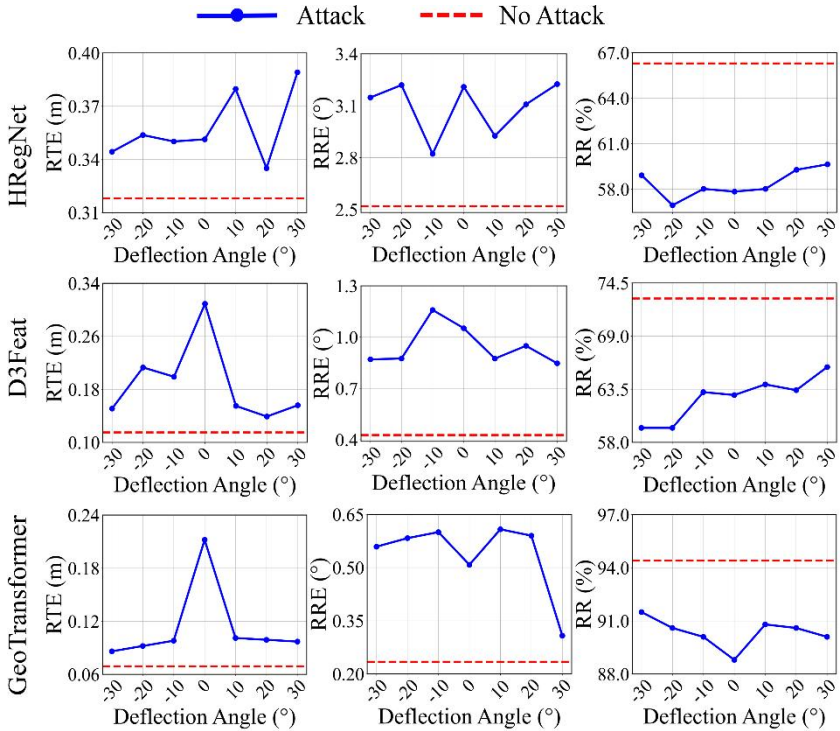

**Fig. 10:** Attack robustness to variations in orientation of ambush tools to target models' performance. Attack parameters are set: K = 5, L = 2.1 m.

- Monitor for roadside occlusions:

An attacker only needs to hang a low-visibility infrared-absorbing sheet alongside the road—no hardware tampering is required—to completely block critical point-cloud regions when a target vehicle passes (in section 5.3). *Original equipment manufacturers (OEMs) and fleet operators should be vigilant for suspicious objects hanging or placed near the roadway during routine inspections. At the software level, they should implement preprocessing checks that detect sudden "disappearances" of point clouds in high-priority regions, raising an alert if these occlusions persist.*

- Account for angular tolerance in testing:

As we demonstrated in section 6.1, the attack remains effective even with a ±30° angular misalignment of the infrared-absorbing material. *Therefore, manufacturers must consider this tolerance into their validation protocols. Testing protocols should include scenarios where occluding objects are placed at various angles—within and beyond ±30°—to ensure that stealth occlusions at any angle do not compromise safety.*

**6.3 Defense discussion**

In this section, we summarize two defense strategies that we refer to as data-level defense, sensor-level defense.

**6.3.1 Data-level defense**

One way to make the model more robust against adversarial attacks is to expose it to adversarial examples during training, which is termed adversarial training. We selected samples ranging from 200 to 1000 from the training set to conduct an adversarial attack. Subsequently, we fed the attacked samples into the models and trained them based on the fine-tuning. Then, we carried out tests on the route 08, 09, and 10.

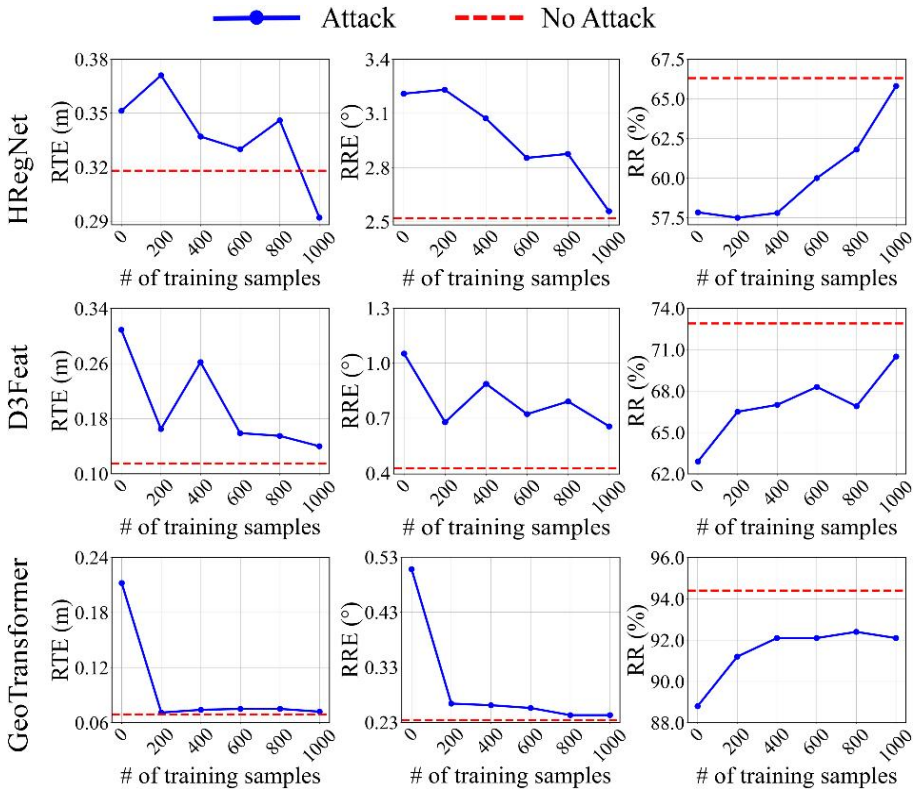

**Fig. 11:** Impact of adversarial training with varying numbers of adversarial samples (ranging from 200 to 1000) on model performance. Adversarial training samples are randomly selected from all post-attack samples. The attack parameters are set to K = 5 and L = 2.1 m.

For HRegNet model, Fig. 11 shows RTE and RRE show a rapid downward trend, while RR rises rapidly. When trained with 1000 samples, the RTE is unexpectedly lower than that without region-hiding attack, and the RRE and RR nearly approach the baseline level without region hidden attacks, demonstrating its anti-attack performance.

For D3Feat model, RTE and RRE fluctuate with a decreasing trend. When 1000 training samples are used, RR is close to the results before attack, indicating that this model could gradually recover its performance under adversarial sample training.

For GeoTransformer model, when 200 adversarial samples are used, the RTE and RRE dropped to levels close to pre-attack results, and the RR increased significantly. However, as the number of adversarial samples increases, the changes in RTE, RRE, and RR are all relatively small, suggesting that the performance of this model tends to stabilize after reaching a certain number of training samples.

Overall, through the study of the training effects of different numbers of adversarial samples, it is found that adversarial sample training can significantly improve the robustness of models against region-hiding attack. Although different models show different performances during the training process, they can all resist region-hiding attacks to a certain extent and reduce the impact of attacks on model performance. This indicates that adversarial sample training is an effective defense strategy that can be used in practical applications to enhance the safety and reliability of LiDAR-based localization module of AV systems.

#### 6.3.2 Sensor-level defense

Although our attack method utilize the black fabric with near-infrared absorption properties to occlude key regions in LiDAR point clouds, rendering them difficult to detect by LiDAR sensors, this approach has a critical limitation: the fabric remains visible in visual images, particularly when deployed in unexpected locations (*e.g.*, large patches on roadside verges). To leverage this vulnerability and enhance overall system security, we propose a multi-sensor fusion-based anomaly detection framework:

- Vision-LiDAR Consistency Check:

By leveraging the extrinsic calibration between the front-facing camera and LiDAR, images can be back-projected into the point cloud. If there is a suspicious object (*e.g.*, black fabric) detected in the image while the corresponding area in the LiDAR data shows low point density or a complete void, it is identified as a potential anomaly.

- Semantic-Based Anomaly Region Perception:

We can train a lightweight semantic segmentation network to identify typical structures along roadways (*e.g.*, trees, utility poles, traffic infrastructure) within the scene. When black fabric is hung on temporary frames, its structural and color features fall into low-confidence or unknown categories in the semantic model. Combined with positional features, such objects can be marked as suspicious objects. The autonomous driving system can alert passengers to potential threats.

- Auxiliary Spectral Information:

In typical scenarios where visible-light cameras perform poorly, such as in strong backlighting, low-light conditions, or under intense glare, the image quality of optical cameras deteriorates significantly, making it difficult to effectively identify targets. In such cases, although black fabrics may exhibit obvious thermal signatures during the day due to their strong heat absorption, thermal imaging can still provide valuable perception capabilities in low-light or no-light environments. At the same time, millimeter-wave radar is sensitive to low-reflectivity materials and can detect the presence of targets in environments with poor visibility or complex lighting conditions. Therefore, in scenarios where camera performance is limited, greater reliance should be placed on thermal imaging and millimeter-wave radar as the primary sensing modalities. By employing a multi-sensor fusion strategy, the system's target recognition capability under complex lighting conditions can be significantly enhanced.

### 6.3 Limitation and future work

Our key region removal attack strategy has several limitations:

- For keypoint-free models like GeoTransformer, if attackers hide too many key regions, they may achieve the attack in some scenarios, but it is too noticeable. If attackers only cover a few key regions, it is hard to achieve the attack goal.

The constraints on the locations of the extracted keypoints, if all keypoints are located in areas that are inaccessible to humans, it will be impossible to carry out field deployment.

- Obtaining keypoints for each point cloud registration model, respectively is quite demanding and intricate. Every registration model varies in feature point extraction mechanisms and matching algorithms, which means that different registration models may produce keypoints at different locations. Consequently, in field deployment, we need to deploy the selected positions extracted by each model, separately, which is quite labor-intensive and time-consuming.

Therefore, we will propose a model that can select universally high-contribution keypoints among different registration models, which can greatly simplify the laborious task of correspondence extraction and reduce the workload of the field deployment.

### 7. Conclusion

This paper studies the impact of key regions removal attack on deep-learning-based LiDAR-based localization system. Extensive evaluations show that our Top-K attack consistently outperforms other schemes such as Rand-K and Min-K attack scheme. For HRegNet, removing top k correspondences with the highest contributions led to a significant degradation in its registration performance. For D3Feat, removing Top-K key-points achieves a comparable adverse effect. Even for the so-called keypoint-free network GeoTransformer, its registration performance dropped sharply when a reasonable number of Top-K superpoints are removed. We also conducted a region removal attack on the localization module (using the NDT algorithm) of the Autoware platform. Despite the support of GPS and IMU, our attack remained effective. Based on our findings, we also carry out a physically realizable attack. The physical experiment results show that deploying some near-infrared absorbing cloth in front of key regions could achieve our attack objective. Finally, we proposed two effective defense strategies to mitigate the threat. Among them, we implemented a data-driven defense strategy. Training different numbers of adversarial samples could enhance the model's defense effect against the DisorientLiDAR attack. Overall, our study filled a research gap in the security of point cloud registration for autonomous vehicles.

**Acknowledge**

This work is supported by the National Key R&D Program of China (No.2022ZD0119000), Hunan Provincial Key R&D Program of China (No.2024JK2020 and 2024JK2021), Hunan Provincial Natural Science Foundation of China (No.2024JJ10027), Young Talents of Huxiang (No.Z202433000575), and Changsha Science Fund for Distinguished Young Scholars (kq2306002).

**Conflict of Interest**
There is no conflict of interest.

**Supporting Information**
Not applicable.

**CRediT Statement**
Yizhen Lao: Conceptualization, Methodology, Writing original draft.
Yu Zhang: Data curation, Experiment, Visualization.
Ziting Wang: Investigation, Resources.
Chengbo Wang: Validation, Review & editing.
Yifei Xue: Validation, Review & editing.
Wanpeng Shao: Validation, Project administration.

**References**

[1] F. Lu, G. Chen, Y. Liu, L. Zhang, S. Qu, S. Liu, R. Gu, C. Jiang, HRegNet: A Hierarchical Network for Efficient and Accurate Outdoor LiDAR Point Cloud Registration, *IEEE Transactions on Pattern Analysis and Machine Intelligence,* 2023, **45**, 11884–11897, doi: 10.1109/TPAMI.2023.3284896.

[2] X. Bai, Z. Luo, L. Zhou, H. Fu, L. Quan, C.-L. Tai, D3Feat: Joint Learning of Dense Detection and Description of 3D Local Features, *Proceedings of the IEEE/CVF Conference on Computer Vision and Pattern Recognition (CVPR)*, 2020, 6358-6366, doi: 10.1109/CVPR42600.2020.00639.

[3] Z. Qin, H. Yu, C. Wang, Y. Guo, Y. Peng, S. Ilic, D. Hu, K. Xu, GeoTransformer: Fast and Robust Point Cloud Registration with Geometric Transformer, *IEEE Transactions on Pattern Analysis and Machine Intelligence*, 2023, **45**, 9806–9821, doi: 10.1109/TPAMI.2023.3259038.

[4] A. Geiger, P. Lenz, R. Urtasun, Are we ready for autonomous driving? The KITTI vision benchmark suite, *2012 IEEE Conference on Computer Vision and Pattern Recognition (CVPR)*, 2012, 3354–3361, doi: 10.1109/CVPR.2012.6248074.

[5] F. Pomerleau, M. Liu, F. Colas, R. Siegwart, Challenging data sets for point cloud registration algorithms, *The International Journal of Robotics Research*, 2012, **31**, 1705-1711, doi: 10.1177/0278364912458814.

[6] H. Caesar, V. Bankiti, A. H. Lang, S. Vora, V. E. Liong, Q. Xu, A. Krishnan, Y. Pan, G. Baldan, O. Beijbom, nuScenes: A multimodal dataset for autonomous driving, *2020 IEEE/CVF Conference on Computer Vision and Pattern Recognition (CVPR)*, 2020, 11618–11628, doi: 10.1109/CVPR42600.2020.01164.

[7] A. Kloukiniotis, A. Papandreou, C. Anagnostopoulos, A. Lalos, P. Kapsalas, D.-V. Nguyen, K. Moustakas, CarlaScenes: A synthetic dataset for odometry in autonomous driving, *2022 IEEE/CVF Conference on Computer Vision and Pattern Recognition Workshops (CVPRW)*, 2022, 4519–4527, doi: 10.1109/CVPRW56347.2022.00498.

[8] Y. Cao, J. Ma, K. Fu, S. Rampazzi, M. Mao, Demo: Automated tracking system for LiDAR spoofing attacks on moving targets, *Third International Workshop on Automotive and Autonomous Vehicle Security*, 2021, doi: 10.14722/autosec.2021.23019

[9] Y. Cao, C. Xiao, B. Cyr, Y. Zhou, W. Park, S. Rampazzi, Q. A. Chen, K. Fu, Z. M. Mao, Adversarial sensor attack on LiDAR-based perception in autonomous driving, *Proceedings of the 2019 ACM SIGSAC Conference on Computer and Communications Security (CCS '19)*, 2019, 2267–2281, doi: 10.1145/3319535.3339815.

[10] H. Shin, D. Kim, Y. Kwon, Y. Kim, Illusion and dazzle: adversarial optical channel exploits against LiDARs for automotive applications, *Proceedings of the 19th International Conference on Cryptographic Hardware and Embedded Systems (CHES 2017)*, 2017, **10529**, 445–467, doi: 10.1007/978-3-319-66787-4_22.

[11] J. Sun, Y. Cao, Q. A. Chen, Z. M. Mao, Towards robust LiDAR-based perception in autonomous driving: general black-box adversarial sensor attack and countermeasures, *Proceedings of the 29th USENIX Security Symposium (USENIX Security '20)*, 2020, 877--894, https://www.usenix.org/conference/usenixsecurity20/presentation/sun.

[12] Y. Cao, N. Wang, C. Xiao, D. Yang, J. Fang, R. Yang, Q. Chen, M. Liu, B. Li, Invisible for both camera and LiDAR: security of multi-sensor fusion based perception in autonomous driving under physical-world attacks, *Proceedings of the 42nd IEEE Symposium on Security and Privacy (IEEE S&P '21)*, 2021, 176–194, doi: 10.1109/SP40001.2021.00076.

[13] Z. Hau, K. Co, S. Demetriou, E. Lupu, Object removal attacks on LiDAR-based 3D object detectors, *Proceedings of the 2021 Workshop on Automotive and Autonomous Vehicle Security (AutoSec)*, 2021, 17–23, doi: https://dx.doi.org/10.14722/autosec.2021.23016.

[14] J. Tu, M. Ren, S. Manivasagam, M. Liang, B. Yang, R. Du, F. Cheng, R. Urtasun, Physically realizable adversarial examples for LiDAR object detection, *Proceedings of the IEEE/CVF Conference on Computer Vision and Pattern Recognition (CVPR)*, 2020, 13713–13722, doi: 10.1109/CVPR42600.2020.01373.

[15] Y. Zhu, C. Miao, T. Zheng, F. Hajiajaghajani, L. Su, C. Qiao, Can we use arbitrary objects to attack LiDAR perception in autonomous driving?, *Proceedings of the 28th ACM SIGSAC Conference on Computer and Communications Security (CCS '21)*, 2021, 1945–1960, doi: 10.1145/3460120.3484738.

[16] Y. Cao, C. Xiao, A. Anandkumar, D. Xu, M. Pavone, AdvDO: realistic adversarial attacks for trajectory prediction, *Proceedings of the 17th European Conference on Computer Vision (ECCV 2022)*, 2022, **13665**, 36–52, doi: https://doi.org/10.1007/978-3-031-20065-6_3.

[17] M. Fukunaga, T. Sugawara, Random Spoofing Attack against LiDAR-Based Scan Matching SLAM, *Symposium on Vehicles Security and Privacy (VehicleSec 2024)*, 2024, doi: 10.14722/vehiclesec.2024.23014.

[18] M. Magnusson, A. Lilienthal, T. Duckett, Scan registration for autonomous mining vehicles using 3D-NDT, *Journal of Field Robotics*, 2007, **24**, 803–827, doi: 10.1002/rob.20204.

[19] N. Mellado, N. Mitra, D. Aiger, Super 4PCS fast global pointcloud registration via smart indexing, *Computer Graphics Forum*, 2014, **33**, 205-215, doi: 10.1111/cgf.12446.

[20] R. B. Rusu, N. Blodow, M. Beetz, Fast point feature histograms (FPFH) for 3D registration, 2009 *IEEE International Conference on Robotics and Automation*, 2009, 3212–3217, doi: 10.1109/ROBOT.2009.5152473.

[21] I. Vizzo, T. Guadagnino, B. Mersch, L. Wiesmann, J. Behley, C. Stachniss, KISS-ICP: in defense of point-to-point ICP simple, accurate, and robust registration if done the right way, *IEEE Robotics and Automation Letters*, 2023, **8**, 1029-1036, doi: 10.1109/LRA.2023.3236571.

[22] C. R. Qi, H. Su, K. Mo, L. J. Guibas, PointNet: deep learning on point sets for 3D classification and segmentation, *IEEE Conference on Computer Vision and Pattern Recognition (CVPR)*, 2017, 77–85, doi: 10.1109/CVPR.2017.16.

[23] Y. Aoki, H. Goforth, A. S. Rangaprasad, S. Lucey, PointNetLK: robust & efficient point cloud registration using PointNet, *IEEE Conference on Computer Vision and Pattern Recognition (CVPR)*, 2019, 7156–7165, doi: 10.1109/CVPR.2019.00733.

[24] C. Choy, W. Dong, V. Koltun, Deep global registration, *IEEE Conference on Computer Vision and Pattern Recognition (CVPR)*, 2020, 2511–2520, doi: 10.1109/CVPR42600.2020.00259.

[25] C. Shi, X. Chen, H. Lu, W. Deng, J. Xiao, B. Dai, RDMNet: reliable dense matching based point cloud registration for autonomous driving, *IEEE Transactions on Intelligent Transportation Systems*, 2023, **24**, 11372–11383, doi: 10.1109/TITS.2023.3286464.

[26] I. Hroob, B. Mersch, C. Stachniss, M. Hanheide, Generalizable stable points segmentation for 3D LiDAR scan-to-map long-term localization, *IEEE Robotics and Automation Letters*, 2024, **9**, 3546–3553, doi: 10.1109/LRA.2024.3368236.

[27] K. Dai *et al.*, LiDAR-based sensor fusion SLAM and localization for autonomous driving vehicles in complex scenarios, *Journal of Imaging (J Imaging)*, 2023, **9, 52**, doi: https://doi.org/10.3390/jimaging9020052.

[28] S. Srinara, Y.-T. Chiu, J.-A. Chen, K.-W. Chiang, M.-L. Tsai, N. El-Sheimy, Strategy on high-definition point cloud map creation for autonomous driving in highway environments, *The International Archives of the Photogrammetry, Remote Sensing and Spatial Information Sciences*, 2023, XLVIII-1, 849–854, doi: 10.5194/isprs-archives-XLVIII-1-W2-2023-849-2023.

[29] J. Cui, W. Jia, J. Zhang, J. Sun, B. Chen, L. Li, Real-time generation and automatic update of 3D point cloud maps in featureless environments based on multi-sensor fusion, *2023 IEEE International Conference on Unmanned Systems (ICUS)*, 2023, 677–682, doi: 10.1109/ICUS58632.2023.10318508.

[30] H. Naderi, I. V. Bajić, Adversarial attacks and defenses on 3D point cloud classification: a survey, *IEEE Access*, 2023, **11**, 144274–144295, doi: 10.1109/ACCESS.2023.3345000.

[31] Y. Cao, S. H. Bhupathiraju, P. Naghavi, T. Sugawara, Z. M. Mao, S. Rampazzi, You can't see me: physical removal attacks on LiDAR-based autonomous vehicles driving frameworks, *32nd USENIX Security Symposium (USENIX Security 23)*, 2023, 2993–3010, https://www.usenix.org/conference/usenixsecurity23/presentation/cao.

[32] Z. Jin, X. Ji, Y. Cheng, B. Yang, C. Yan, W. Xu, Pla-lidar: physical laser attacks against lidar-based 3D object detection in autonomous vehicle, *2023 IEEE Symposium on Security and Privacy (SP)*, 2023, 1822–1839, doi: 10.1109/SP46215.2023.10179458.

[33] T. Sato, Y. Hayakawa, R. Suzuki, Y. Shiiki, K. Yoshioka, Q. A. Chen, WIP: practical removal attacks on LiDAR-based object detection in autonomous driving, *ISOC Symposium on Vehicle Security and Privacy (VehicleSec),* 2023, doi: https://dx.doi.org/10.14722/vehiclesec.2023.23036.

[34] J. Sun, Y. Cao, Q. A. Chen, Z. M. Mao, Towards robust LiDAR-based perception in autonomous driving: general black-box adversarial sensor attack and countermeasures, *29th USENIX Security Symposium (USENIX Security 20)*, 2020, 877–894, https://www.usenix.org/conference/usenixsecurity20/presentation/sun.

[35] R. Spencer Hallyburton and Yupei Liu and Yulong Cao and Z. Morley Mao and Miroslav Pajic, Security Analysis of Camera-LiDAR Fusion Against Black-Box Attacks on Autonomous Vehicles, *31st USENIX Security Symposium (USENIX Security 22)*, 2022, 1903–1920, https://www.usenix.org/conference/usenixsecurity22/presentation/hallyburton.

[36] J. Tu, M. Ren, S. Manivasagam, M. Liang, B. Yang, R. Du, F. Cheng, R. Urtasun, Physically realizable adversarial examples for LiDAR object detection, *IEEE Conference on Computer Vision and Pattern Recognition (CVPR)*, 2020, 13713-13722, doi: 10.1109/CVPR42600.2020.01373.

[37] Y. Zhu, C. Miao, T. Zheng, F. Hajiajaghajani, L. Su, C. Qiao, Can we use arbitrary objects to attack LiDAR perception in autonomous driving?, *Proceedings of the 2021 ACM SIGSAC Conference on Computer and Communications Security (CCS '21)*, 2021, 1945–1960, doi: https://doi.org/10.1145/3460120.3485377.

[38] Z. Xiang, D. J. Miller, S. Chen, X. Li, G. Kesidis, A backdoor attack against 3D point cloud classifiers, *IEEE/CVF International Conference on Computer Vision (ICCV),* 2021, 7577-7587, doi: 10.1109/ICCV48922.2021.00750.

[39] Y. Lou, Y. Zhu, Q. Song, R. Tan, C. Qiao, W.-B. Lee, J. Wang, A First Physical-World Trajectory Prediction Attack via LiDAR-induced Deceptions in Autonomous Driving, *33rd USENIX Security Symposium (USENIX Security 24)*, 2024, 6291–6308, https://www.usenix.org/conference/usenixsecurity24/presentation/lou.

[40] P. Kumar, D. Vattikonda, K. Bhat, P. Kalra, SLACK: attacking LiDAR-based SLAM with adversarial point injections, *2024 IEEE International Conference on Image Processing Challenges and Workshops (ICIPCW)*, 2024, 4082–4088, doi: 10.1109/ICIPCW64161.2024.10769168.

[41] K. Yoshida, M. Hojo, T. Fujino, Adversarial scan attack against scan matching algorithm for pose estimation in LiDAR-based SLAM, *IEICE Transactions on Fundamentals of Electronics, Communications and Computer Sciences*, 2022, E105-A, 326–335, doi: 10.1587/transfun.2021CIP0017.

[42] P. J. Besl, N. D. McKay, A method for registration of 3-D shapes, *IEEE Transactions on Pattern Analysis and Machine Intelligence*, 1992, **14**, 239–256, doi: 10.1109/34.121791.

[43] W. Zhang, J. Qi, P. Wan, H. Wang, D. Xie, X. Wang, G. Yan, An easy-to-use airborne LiDAR data filtering method based on cloth simulation, *Remote Sensing,* 2016, **8**, 501, doi: https://doi.org/10.3390/rs8060501.

[44] T. Shan, B. Englot, D. Meyers, W. Wang, C. Ratti, D. Rus, LIO-SAM: tightly-coupled LiDAR inertial odometry via smoothing and mapping, *IEEE/RSJ International Conference on Intelligent Robots and Systems (IROS)*, 2020, 5135–5142, doi: 10.1109/IROS45743.2020.9341176.

(a) https://www.the-black-market.com/marketplace/vl-flock sheet/
(b) https://github.com/ispc-lab/HRegNet
(c) https://github.com/XuyangBai/D3Feat
(d) https://github.com/qinzheng93/GeoTransformer
(e) https://github.com/autowarefoundation/autoware
(f) http://www.velodynelidar.com

**Publisher's Note:** Engineered Science Publisher remains neutral with regard to jurisdictional claims in published maps and institutional affiliations.